%% file: main.tex
\documentclass[letterpaper, 10 pt, conference]{ieeeconf}  

\IEEEoverridecommandlockouts                              

\DeclareMathAlphabet{\mathcal}{OMS}{cmsy}{m}{n}

\usepackage{amsmath} 
\usepackage{amsfonts}  
\usepackage{booktabs}
\usepackage{graphics} 
\usepackage{graphicx} 
\usepackage{epsfig} 
\usepackage{bm} 
\usepackage{times} 
\usepackage{amssymb}  
\usepackage{bm}
\usepackage{algorithm}
\usepackage{algpseudocode}
\usepackage{mathtools}
\usepackage{listings}
\usepackage{subcaption}
\usepackage{multirow}
\usepackage{hyperref}
\usepackage[table,xcdraw]{xcolor}
\usepackage{optidef}
\usepackage{subcaption}
\usepackage{array}
\usepackage[table]{xcolor}

\begin{document}

\title{\LARGE \bf
Friction-Aware Safety Locomotion for Wheeled-legged Robots using Vision Language Models and Reinforcement Learning}

\author{Bo Peng$^{1*}$, Donghoon Baek$^{2}$, Qijie Wang$^{3}$, and Joao Ramos$^{1,2}$
\thanks{This work is supported by the National Science Foundation via grant IIS-2024775.}
\thanks{The authors are with the $^1$Department of Electrical and Computer Engineering and the $^2$Department of Mechanical Science Engineering at the University of Illinois at Urbana-Champaign and the $^3$School of Software at Tsinghua University. *Corresponding author: {
 percypeng5221@gmail.com}}}

\maketitle
\thispagestyle{empty}
\pagestyle{empty}

\begin{abstract}
Controlling Wheeled-legged robots is challenging especially on slippery surfaces due to their dependence on continuous ground contact. Unlike quadrupeds or bipeds, which can leverage multiple fixed contact points for recovery, wheeled-legged robots are highly susceptible to slip, where even momentary loss of traction can result in irrecoverable instability. Anticipating ground physical properties such as friction before contact would allow proactive control adjustments, reducing slip risk. In this paper, we propose a friction-aware safety locomotion framework that integrates Vision-Language Models (VLMs) with a Reinforcement Learning (RL) policy. Our method employs a Retrieval-Augmented Generation (RAG) approach to estimate the Coefficient of Friction (CoF), which is then explicitly incorporated into the RL policy. This enables the robot to adapt its speed based on predicted friction conditions before contact. The framework is validated through experiments in both simulation and on a physical customized Wheeled Inverted Pendulum (WIP). Experimental results show that our approach successfully completes trajectory tracking tasks on slippery surfaces, whereas baseline methods relying solely on proprioceptive feedback fail. These findings highlight the importance and effectiveness of explicitly predicting and utilizing ground friction information for safe locomotion. They also point to a promising research direction in exploring the use of VLMs for estimating ground conditions, which remains a significant challenge for purely vision-based methods.

\end{abstract}

\input{1_introduction}

\input{2_background}
\input{3_methods}
\input{4_experiments}

\input{5_results_and_analysis}

\input{6_discussion_and_limitations}
\input{7_conclusion}



\bibliographystyle{IEEEtran}
\bibliography{main}


\end{document}

%% file: 1_introduction.tex
\section{Introduction}

Wheeled-legged robots combine the speed of wheels with the adaptability of legs, making them ideal for applications from industrial automation to disaster response \cite{purushottam2022hands, klemm2020lqr, purushottam2023dynamic}. These systems are particularly vulnerable when traversing slippery terrain compared to quadrupedal or bipedal robots, primarily due to their reliance on continuous rolling contact with the ground. Once slipping occurs, the wheels can lose effective traction and continue spinning without generating meaningful ground reaction forces. This loss of contact makes it extremely difficult for the system to stabilize or recover from disturbances. Such robots would greatly benefit from the ability to predict environmental conditions like slippery surfaces or foot sinkage through exteroceptive sensing before making contact.

Most control strategies for wheeled-legged robots rely on direct interaction with the terrain, requiring physical contact and adaptation through proprioceptive feedback. Although both model-based approaches \cite{klemm2020lqr, baek2024real} and reinforcement learning (RL)-based locomotion controllers \cite{agarwal2023legged, Lee_2020, hwangbo2019learning, loquercio2022learning, kumar2021rma, kumar2022adapting} have demonstrated the ability to handle complex terrains in legged systems, they cannot be directly applied to wheeled-legged platforms due to their distinct contact dynamics. Hybrid approaches that combine model-based and learning-based methods have been proposed to leverage the advantages of both and mitigate their respective limitations \cite{jenelten2024dtc, kang2023rl+, baek2022hybrid}. However, as all of these methods depend solely on proprioceptive sensing, they require reactive recovery strategies to adapt the robot’s behavior after a slip occurs, limiting their effectiveness in preventing falls proactively.

\begin{figure}[t]
  \centering
\includegraphics[width=0.9\linewidth]{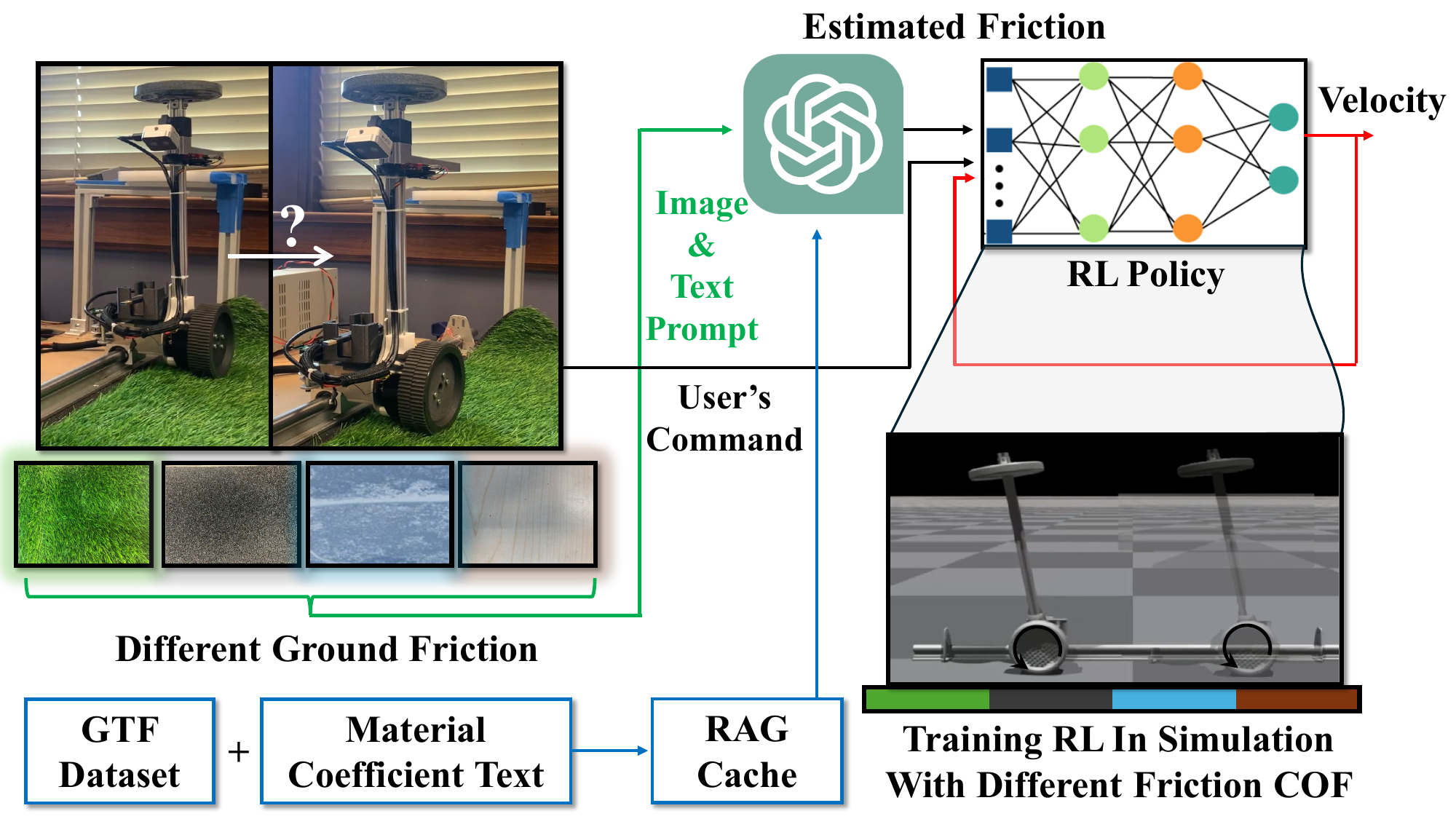}
  \caption{\textbf{Overview of Friction-Aware Safety Locomotion Framework using Vision Language Model.} This shows how a Vision Language Model integrates with an RL policy to control a wheeled robot on slippery surfaces. The Friction-from-Vision (FFV) module estimates ground friction using image and text data, enabling the RL policy to adapt behavior based on friction coefficients before contact. Developed in simulation, the pre-trained policy transfers successfully to the real world without manual tuning.}
  \label{fig:concept_img}
  \vspace{-2em}
\end{figure} 

One promising direction is to estimate ground properties, such as friction, using vision-based methods. However, friction inherently depends on contact interactions and force transmission, making it difficult to estimate accurately from visual information alone. Learning friction estimation in simulation is also challenging since replicating the physical effects of terrain interaction (e.g., stiffness and friction) remains difficult. Friction-From-Vision (FFV) is not a new topic, but a generalizable solution and a large-scale dataset are still lacking. Collecting ground-truth friction coefficients typically requires specialized physical instrumentation, which poses a significant barrier to building large-scale, labeled datasets.

To enable safer locomotion of wheeled-legged robots on slippery surfaces, we propose a friction-aware control framework that integrates a VLM with RL. Using prompt engineering—such as material descriptions—the VLM infers the ground CoF, allowing the robot to proactively adapt its behavior based on anticipated terrain conditions. This approach addresses the limitations of purely vision-based methods, which lack the semantic understanding required to reason about friction. For instance, identifying a banana peel as slippery relies on commonsense knowledge, not just visual cues. By employing the Retrieval-Augmented Generation (RAG) technique, our method estimates CoF without requiring paired image-friction datasets or collecting ground-truth labels. This eliminates the need to train a new estimator from scratch and enables efficient, scalable friction inference. 

The contributions of this paper are threefold. First, we propose a friction-aware RL locomotion framework that integrates a VLM for the control of wheeled humanoid robots. Second, we introduce a vision-based friction estimation module that utilizes the RAG technique in conjunction with a VLM to estimate ground friction coefficients without requiring the training of a new neural network or collecting paired datasets. Third, we demonstrate the effectiveness of the proposed framework through hardware experiments on a physical WIP, a reduced-order model of a wheeled-legged robot, showing improved safety and tracking performance compared to classical controllers and proprioception-based RL policies.

%% file: 2_background.tex
\section{Related Works}
\label{sec:background}
\subsection{Friction from Vision.} 
Estimating ground friction remains a challenge without a universal solution. Variations in image intensity suggest rough surfaces, correlating with higher friction. Humans often assess slipperiness through visual cues like surface gloss and roughness, though these perceptions can be imprecise \cite{Joh2006WhyWS, lesch2008visually}.  Some methods predict friction by identifying surface types and referencing known databases \cite{brandao2016material}, but these are limited by the lack of a comprehensive friction-vision dataset and environmental factors. VLMs excel in visual-language reasoning tasks, often achieving impressive zero-shot and few-shot performance with minimal training. Inspired by \cite{7803311}, which used text mining to infer CoF, we propose leveraging VLMs' strong generalization and reasoning abilities with visual data to estimate CoF across surfaces. 

\subsection{Locomotion with Friction} 
Recent research on quadruped locomotion highlights robustness and agility, focusing on navigating difficult terrains and performing parkour-like tasks \cite{Lee_2020, hwangbo2019learning, miki2022learning, zhuang2023robot, cheng2023extreme}. Similarly, humanoid locomotion has been extensively studied using both model-based and model-free methods \cite{chignoli2021humanoid, stasse2017talos, kuindersma2020recent, castillo2021robust, radosavovic2024real}. 

Anticipating environmental challenges is crucial for robots to avoid accidents and improve task performance. For example, incorporating visual information increased success rates in stair climbing from 40\% to 100\% \cite{loquercio2022learning}, and bipedal robots have been shown to select less slippery paths for safer navigation \cite{brandao2016material}. This highlights the importance of integrating environmental factors into RL to prevent slips and enhance safety. Recent work also addresses locomotion on slippery ground but requires extensive training \cite{chen2024identifying}.

%% file: 3_methods.tex
\section{Methods}
\label{sec:methods}

The proposed ground friction-aware locomotion framework comprises two key components: 1) the FFV module, which uses VLMs to estimate the ground friction coefficient ($f_t$) from images, and 2) an RL policy trained entirely in simulation, then transferred to the real world with zero-shot learning. The procedure of our framework is shown in Fig. \ref{fig:concept_img}.

\subsection{Friction Estimation using Vision Language Models}
\label{method:A}
We introduce the FFV module, which employs a selected VLM to estimate the COF of the ground. We chose \textit{GPT4-o} due to its performance and convenient API. To have a good friction estimation, we need both vision information and a dataset consisting of friction information of common surfaces to refer to. In this situation, several approaches to utilizing VLMs can be considered, ranked in terms of modification to VLMs: pretraining on a specialized dataset, fine-tuning \cite{hu2021lora}, RAG \cite{lewis2020retrieval}, and prompt engineering. While pretraining and fine-tuning yield promising results, they demand large-scale datasets, which are currently lacking. To address this limitation, we employ RAG with an open-source dataset, supplemented by 94 text-based friction coefficient references \cite{Totten2017, Blau2009}. The dataset used is the \textit{Ground-Truth coefficient of Friction dataset} (GTF) \cite{7803311}, comprising 129 images of 43 walkable surfaces. \\

For each walkable surface in the GTF, we utilized a pre-trained CLIP visual encoder to encode the original image $x_i, i \in N$, where $N$ is the total number of surfaces, into features $f_{\text{img}}^i\in \mathbb{R}^{D}, i \in [1,N]$, $D$ represents the dimension of the CLIP features. The format of the text friction coefficient data is similar to:

\begin{lstlisting}[frame=single]
  "Material": "Wrought iron",
  "Static Coefficient of Friction": 0.44,
  "Against Material": "wrought iron"
\end{lstlisting}

For each piece of data, we construct a simple prompt $t_i$ as \text{"[Material] and [Against Material]"}, $i \in M,$ where $M$ is the number of text friction coefficient data entries, and use the pre-trained CLIP text encoder to encode $t_i$ into features $f_{\text{text}}^i\in \mathbb{R}^{D}, i \in [1,M]$. Subsequently, pairs of image feature-image path and text feature-text are cached in a cache file. For the input image $y$, we still use the CLIP visual encoder to encode its features as $f_y$. We read all $f_x$ and $f_{\text{text}}$ from the cache file. Then they are concatenated into two matrices, $F_{\text{img}} \in \mathbb{R}^{N \times D}$ and $F_{\text{text}} \in \mathbb{R}^{M \times D}$. We calculate the cosine similarity between $f_y$ and $F_{\text{text}}$, and between $f_y$ and $F_{\text{img}}$,

\begin{equation}
\label{cosine similarity}
\begin{split}
\text{cosine\_similarity}(f_y, F_{\text{text}}) = \frac{f_y \cdot F_{\text{text}}^T}{\|f_y\| \|F_{\text{text}}\|}\\
\text{cosine\_similarity}(f_y, F_{\text{img}}) = \frac{f_y \cdot F_{\text{img}}^T}{\|f_y\| \|F_{\text{img}}\|}
\end{split}
\end{equation}
takes the top $K$ as cached knowledge from the image and text caches, and input them together with the input prompt into \textit{GPT4-o}. The input prompt defines the format of the text returned by \textit{GPT4-o}, so the estimated CoF value can be obtained using regular matching in the text returned by \textit{GPT4-o}. Each estimation takes around 5 seconds, affected by internet condition and server burden. \\

\subsection{Friction-Aware Reinforcement Learning}
\label{method:B}
Using history of proprioception as an observation in RL policies has become a standard approach in legged robot control \cite{Lee_2020, hwangbo2019learning, miki2022learning, zhuang2023robot, cheng2023extreme}. The underlying assumption is that proprioceptive data contains enough information to replace privileged parameters such as ground friction. The inherent limitation of this method is that it requires the robot to experience new situations for the proprioception history to capture meaningful data. While this can be done in real-time, for systems with limited ground contact points, reacting and recovering is particularly challenging. A good example is to imagine a human wearing wheeled skates, trying to regain balance to avoid falling—such recovery. For this reason, we decided to explicitly incorporate an estimated ground friction coefficient in the RL policy, rather than relying solely on the robot's proprioception. We trained the RL policy separately from a VLM in simulation. To consider the VLM's estimation error, we introduced Gaussian noise and random estimation errors in the estimated value, which was then fed into the RL policy. \\

\noindent\textbf{Observation.} Having more information does not always mean better performance. Some of it may be redundant and sometimes even increases the \textit{reality gap} \cite{tan2018sim}. In a classical approach, slipping in the WIP system is examined in \cite{sorensen2018wheeled}, where the absolute slip $\gamma$ is defined as $\frac{r\dot{\alpha} - \dot{x}}{2\pi r}$. Here, $r$ is a wheel radius, $\dot{\alpha}$ is the angular velocity, and $\dot{x}$ is the linear velocity. The extended equation for calculating friction is $\hat{f} = \mu\text{sign}(\gamma)$ where $\mu$ is the friction coefficient. Inspired by this, we designed our observation space as follows:
\begin{equation}
\begin{split}
    o_t = \{ x_w, \dot{x}_w, \beta, \dot{\beta}, x_{tr}, a_{t-1}, c_t, \hat{\mu} \}
\end{split}
\label{rl_observation}
\end{equation}
where, $x_w$ and $\dot{x}_w$ represent the angular position and velocity of the wheel, while $\beta$ and $\dot{\beta}$ denote the angular position and velocity of the pole joint. The variable $a_{t-1}$ is the previous action, $c_t$ is the user command, and $\hat{\mu}$ is the estimated friction from the VLM. Additionally, $x_{tr}$ refers to the position of the translation joint. It is important to note that $x_{tr}$, a global position, is not accessible in the real world and is not accurate even if visual odometry is used. When there's a sudden change in $x_{tr}$, we cannot tell whether it's caused by slipping or sensor noise, making this information unreliable. But we observed that including $x_{tr}$ in the observation space improves the tracking performance and stability during RL policy training. This is because the $x_{tr}$ provides the RL policy with additional information about slip, given slipping is caused by the disparity between \( x_{tr} \) and \( x_w \). As a training trick, we initially used \( x_{tr} \) and construct the reward function to reduce the disparity between \( x_{tr} \) and \( x_w \) so the policy is trying to learn to "avoid slipping". When the reward value converges and the performance is stable, we substitute the $x_{tr}$ with $x_w$ in the deployment stage. \\     

\noindent\textbf{Action Space.} The action $a_t$ represents the desired velocity for the wheel joint. Empirically, we found that using velocity mode with a built-in PD controller ($k_p = 0, k_d = 3$) leads to faster training convergence and simplifies sim-to-real transfer compared to directly learning the torque. \\

\noindent\textbf{Reward Function.} The reward function is designed considering the five major components as follows. 

\begin{equation}
\begin{split}
    R_t =& \underbrace{1 - \beta_t^2}_{\text{Keep balance}} - \underbrace{0.01 |\dot{x}_t| + 0.005 |\dot{\beta}_t|}_{\text{Penalty for oscillation}} - \underbrace{3 |\dot{x}_t - r \dot{\alpha}_t|}_{\text{Avoid Slipping}} \\
    &+ \underbrace{\frac{0.3 \zeta_t}{e_t + 0.01} a_{j,t}^2}_{\text{Tracking Component}} - \underbrace{(A_l + (A_h - A_l) \zeta_t) a_{j,t}^2}_{\text{Output penalty}}
\end{split}
\label{rl_reward}
\end{equation}
This encourages the system to track the desired velocity while maintaining stability and preventing slip. Since we use a single actuator to balance and track the signal, producing smooth outputs is crucial for effective sim-to-real transfer. The output penalty term is designed to penalize actions that apply excessive force or velocity in the system, ensuring smoother control. The $A_l$ and $A_h$ are two predefined parameters, the lowest output penalty and the highest output penalty, respectively. We observed that incorporating the "avoiding slipping" term significantly enhances tracking performance. \\

\noindent\textbf{Curriculum Learning.} In the reward function (\ref{rl_reward}), the term $\zeta_t$ contributes to curriculum learning and is defined as follows.
\begin{equation}
    \zeta_t = (tanh(\Bar{T}_t/T_{max} * 8 - 3) + 1)/2
\label{rl_curri}
\end{equation}
where $\Bar{T}_t$ is the current mean episode length and $T_{max}$ is the maximum episode length. Initially, the focus of the training process is on learning balancing, gradually transitioning attention towards the tracking task. The $ \tanh $ function ensures that this parameter increases slowly during the early stages of training, allowing the policy to focus on easier tasks.

\begin{figure}[b]
  \centering
\includegraphics[width=0.9\linewidth]{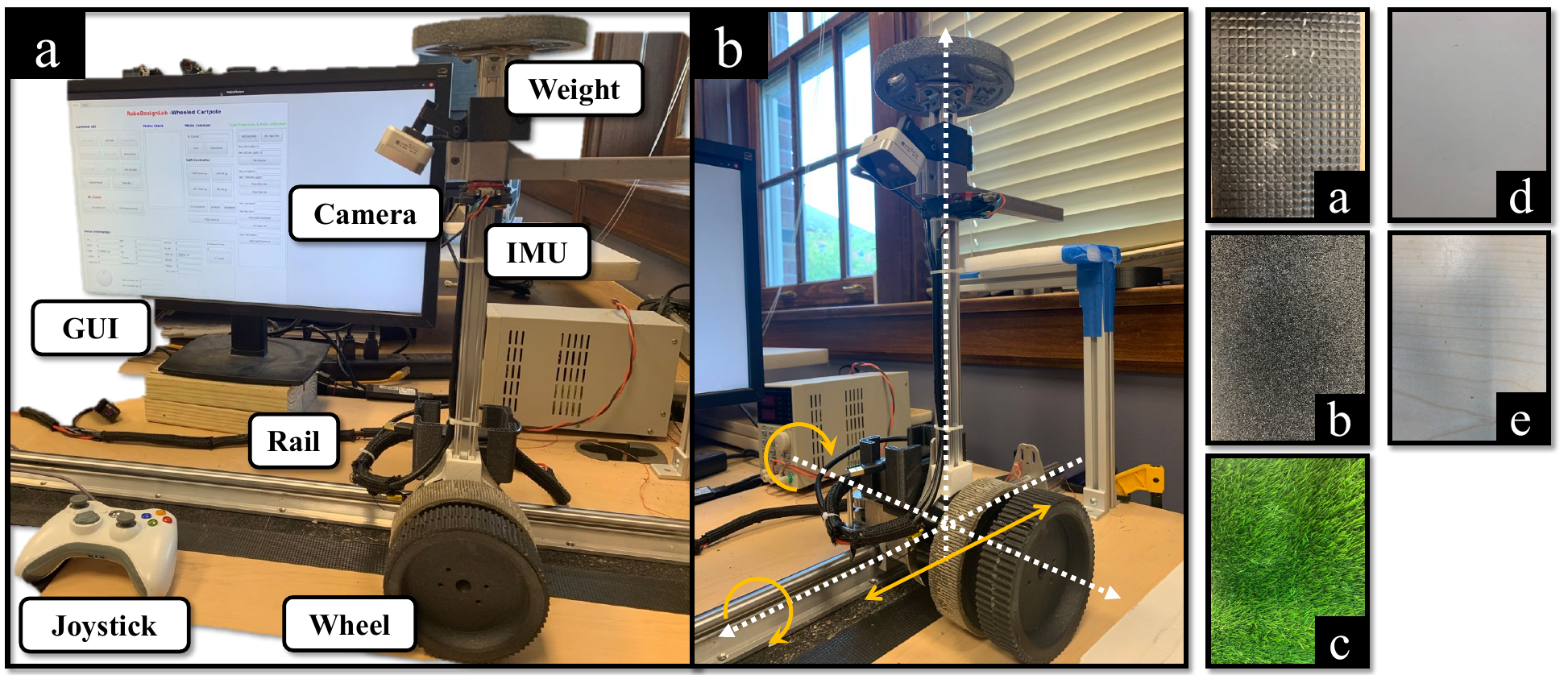}
  \caption{\textbf{The experiments involve testing two types of wheels (smooth and rough) across five surfaces (rubber, anti-slip tape, wood, cardboard, and grass).} The smooth wheel is tested on wood, cardboard, and grass, while the rough wheel is tested on rubber and anti-slip tape. Each wheel-surface combination forms a distinct testing environment. For accurate evaluation, three tests are conducted for each combination, resulting in a total of 15 tests.}
  \label{fig:experimental setup}
\end{figure} 

%% file: 4_experiments.tex
\section{Experiments}
\label{sec:experimental setup}

\subsection{Customized Wheeled-Inverted Pendulum} A customized WIP was developed to verify the proposed framework's feasibility, as shown in Fig. \ref{fig:experimental setup}(a, b). This system is a simplified version of the wheeled-legged robot, SATYRR \cite{purushottam2022hands}. Unlike the traditional inverted pendulum, our WIP system includes a wheel joint and a roll joint, causing slipping to happen more easily and making control more challenging. For example, when the wheel accelerates aggressively, it often lifts off and loses contact with the ground. 

\subsection{Simulation and Hardware details}
\noindent\textbf{Hardware Setup.} The system employs the same actuators as a wheeled humanoids \cite{purushottam2023dynamic}, with an inertial measurement unit (VN-100, VectorNav, USA) mounted on the pole link. We used a RealSense (D405 model) camera to obtain the ground images. Five types of surface materials (rubber, anti-slip tape, wood, cardboard, and grass) and two wheels with different friction conditions are used in the real-world tracking task. The control loop runs at 400 Hz, and the RL policy's decision frequency is 50 Hz. All software is connected via the Robot Operating System (ROS). \\

\noindent\textbf{Simulation Environment Setup.} 
We utilize the IsaacGym simulator to train our RL policy and all baselines for WIP. We configure the static and dynamic friction of the ground to be 0, and we adjust the friction condition of the WIP's wheel accordingly. IsaacGym calculates the CoF between two surfaces as the average of the CoFs of each surface.

\begin{table}[t]
    \centering
    \resizebox{\linewidth}{!}{ 
    \begin{tabular}{ll|ll}
        \toprule
        \textbf{Model} & \textbf{Network} & \textbf{Model} & \textbf{Network} \\
        \midrule
        \multirow{5}{*}{Actor} & FC (10, 64) & \multirow{12}{*}{Adaptation Module} & Conv1d \\
         & Lrelu &  & Length = 40 \\
         & FC (64, 64) &  & Num of Kernels = 64 \\
         & Lrelu &  & Kernel Size = 3 \\
         & FC (64, 1) &  & Stride = 1 \\
         \cmidrule(lr){1-2}
         \multirow{5}{*}{Critic}& FC (10, 64) &  & Padding = 1 \\
         & Lrelu &  & FC(2560, 256) \\
         & FC (64, 64) &  & Relu \\
         & Lrelu &  & FC(256, 64) \\
         & FC (64, 1) &  & Relu \\
         \cmidrule(lr){1-2}
         \multirow{3}{*}{Environment Encoder}& FC (1, 2) &  & FC(64, 8) \\
         & Lrelu &  & Relu \\
         & FC (2, 1) &  & FC(8, 1) \\
        \bottomrule
    \end{tabular}
    }
    \caption{\textbf{Architecture of PPO and RMA.} The symbol $FC(a, b)$ denotes a fully connected layer with input dimension $a$ and output dimension $b$. The term $Lrelu$ refers to the leaky ReLU. We adopted this architecture for both our method and all baseline approaches.}
    \label{tab:model_architecture}
    \vspace{-2em}
\end{table}

\subsection{Experimental Plan}
\noindent The experiments aim to: 1) validate the friction estimation performance of the FFV module; 2) assess how effectively our framework completes the tracking task without failure across various surface types compared to baseline methods; and 3) analyze the impact of incorporating privileged information $x_{tr}$ into the observation space. \\

\noindent\textbf{Ground Friction Coefficient Estimation of FFV Module.} We evaluated the FFV module using the GTF dataset \cite{7803311}, with Root Mean Squared Error (RMSE) as the metric and 2-, 5-, and 10-fold cross-validation following \cite{7803311}. For baselines, we compared against a vision-based method from \cite{brandao2016material} and the Word Material-Material similarity (WordMM) method from \cite{7803311}, representing vision-only and commonsense-based friction estimation, respectively. The former uses a CNN trained on GTF, but due to outdated code, we implemented a Vision Transformer (ViT) \cite{dosovitskiy2020image} following the same training pipeline. \\

\noindent\textbf{Safety Tracking Performance Evaluation.} We conducted safety-tracking tasks in simulation and the real world using our WIP. We prioritize success rate as our primary evaluation criterion while also assessing the tracking performance of our method and baseline approaches. For the baselines, the classical LQR \cite{kalman1960contributions}, PPO \cite{schulman2017proximal}, PPO with Domain Randomization (DR)\cite{tobin2017domain} and Teacher and Student Policy from RMA \cite{kumar2021rma}, which are commonly used to control wheeled robots, are selected. In the observation part, the main difference between ours and PPO is that our method uses CoF explicitly. To ensure fairness of comparison and follow the common trend, we use the DR technique for friction in PPO.

\begin{figure}[t]
  \centering
\includegraphics[width=\linewidth]{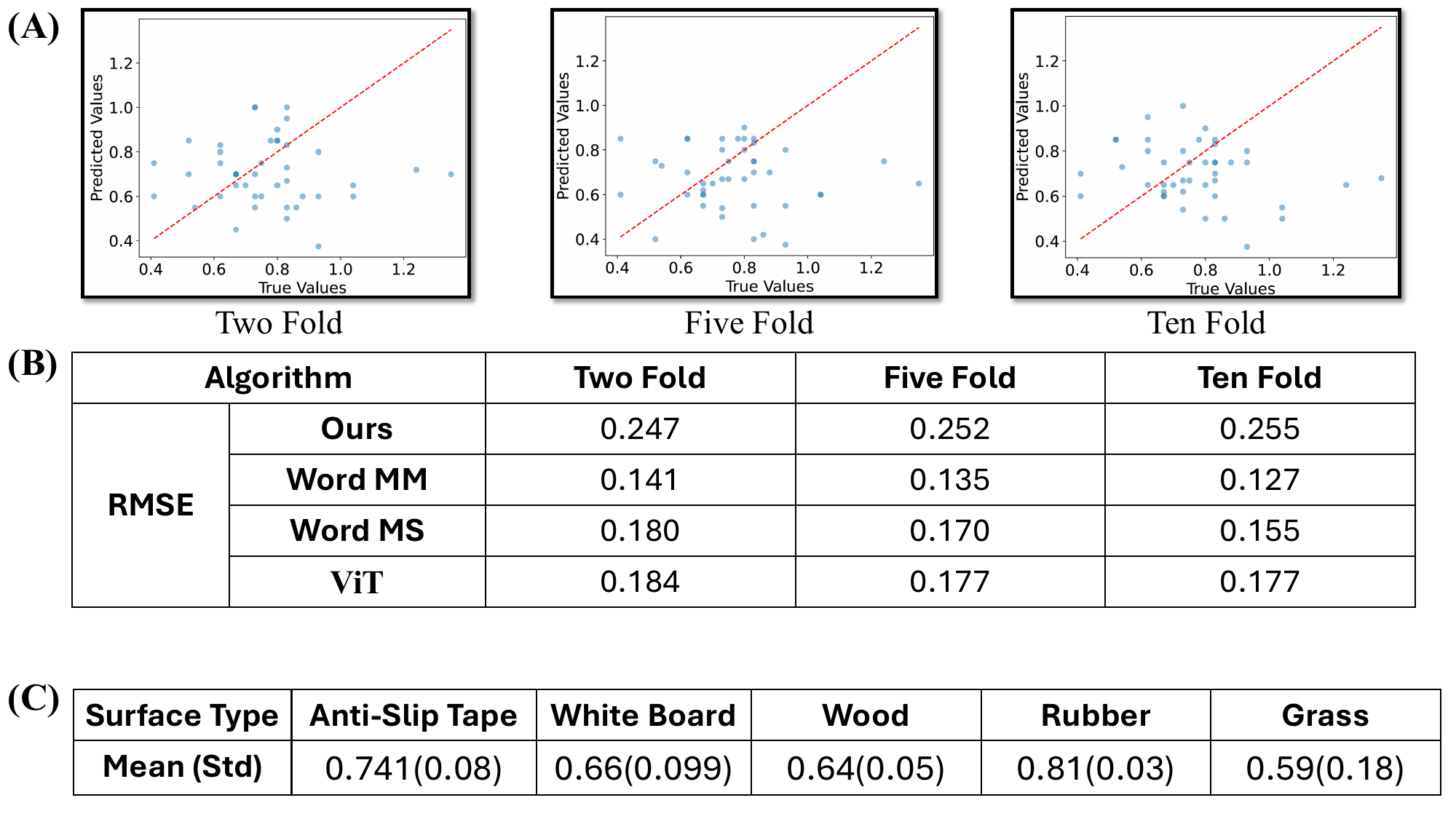}
  \caption{\textbf{Results of Friction Coefficient Estimation.} (A) Cross-validation was employed due to the limited dataset size, with results depicted for 2, 5, and 10-fold validation. (B) Comparison of friction coefficient estimation performance between ours and other methods. For the Word MM, we got the result from the paper. (C) Estimation of friction coefficients for various surface types, presented as the mean and standard deviation of ten trials. Note that although the friction estimation may not be highly precise, it provides a reliable indication of whether the ground is slippery, which is sufficient for adjusting the desired velocity accordingly. Moreover, the FFV module does not require collecting a large dataset or training a neural network from scratch, making it a practical and efficient solution. }
  \label{fig:friction estimation result}
\end{figure}

To implement RMA, the privileged information includes the CoF, with the encoder network generating an extrinsic vector $z_t$ of size 1 based on this coefficient. To ensure a fair comparison, we keep the control frequency, environment setup, and training duration consistent across all methods, training each policy until reward saturation. We select a neural network that achieves as high a reward as possible without exhibiting high-frequency outputs characteristic of bang-bang control, in order to minimize the sim-to-real gap. \\

\begin{figure}[t]
  \centering
\includegraphics[width=\linewidth]{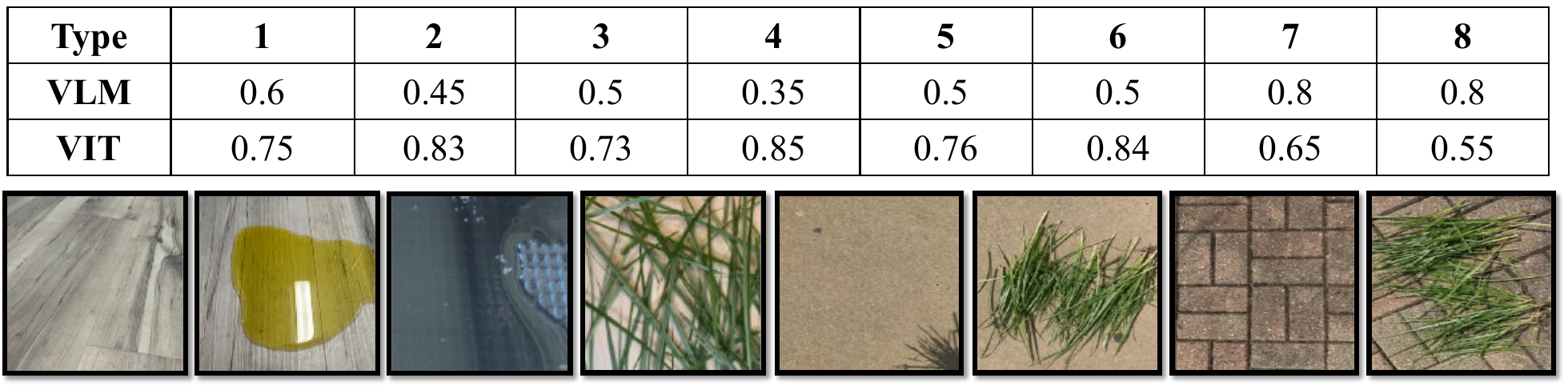}
  \caption{\textbf{Friction coefficient estimation results for images with different textures.} Applying oil, grass, or water to the surface increases slipperiness. \textit{GPT-4o} can recognize these changes and adjust the CoF output accordingly, whereas the Vision Transformer lacks this capability.}
  \label{fig:cof_addition_result}
\end{figure}

\noindent\textbf{Ablation Study on Translation Joint.} 
It is not natural to adopt translation joints as observations because it is difficult to obtain actual values from hardware. As discussed in Section \ref{method:B}, the difference between the velocity of the translation joint and that of the wheel joint is crucial for detecting slip. In the ablation study, we focus on evaluating the impact of incorporating the translation joint on RL tracking performance in slippery conditions. In addition, we also explore how simple Domain Randomization (DR) \cite{tobin2017domain} can contribute to the anti-slip effect.

\begin{figure}[t]
  \centering
    \includegraphics[width=\linewidth]{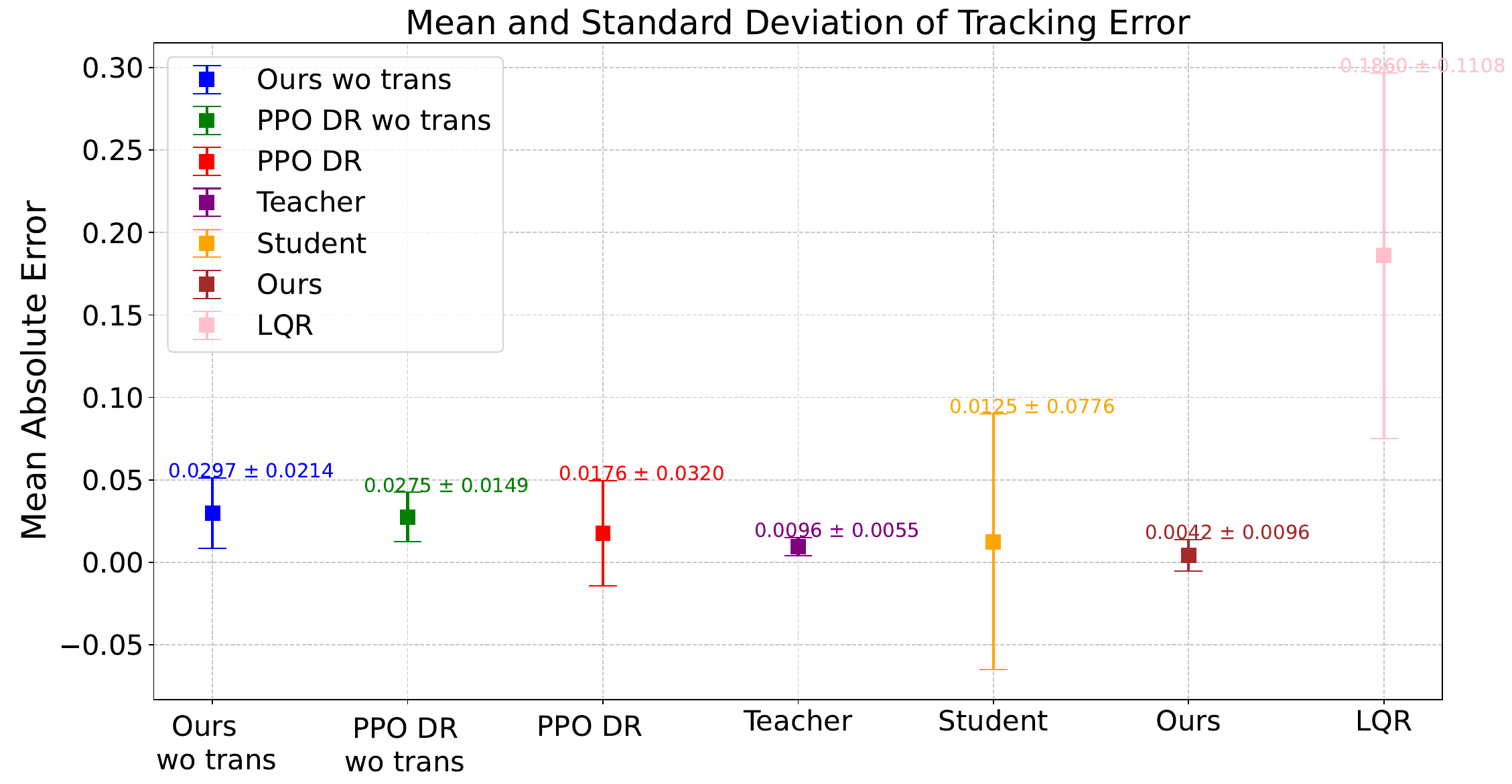}
    \caption{\textbf{Comparison Result of Tracking Task in Simulation.} The system begins at zero position with desired position range of [-0.3m, 0.3m]. A single policy is trained for all terrains and evaluated. The mean and standard deviation are calculated from 1,000 results tested across various ground surface types.}
    \label{fig:tracking_error_comparison}
\end{figure}

\begin{figure}[t]
    \centering
    \setlength{\tabcolsep}{1pt}  
    \renewcommand{\arraystretch}{0.8}  
    \begin{tabular}{ccc}
        \includegraphics[width=0.33\linewidth]{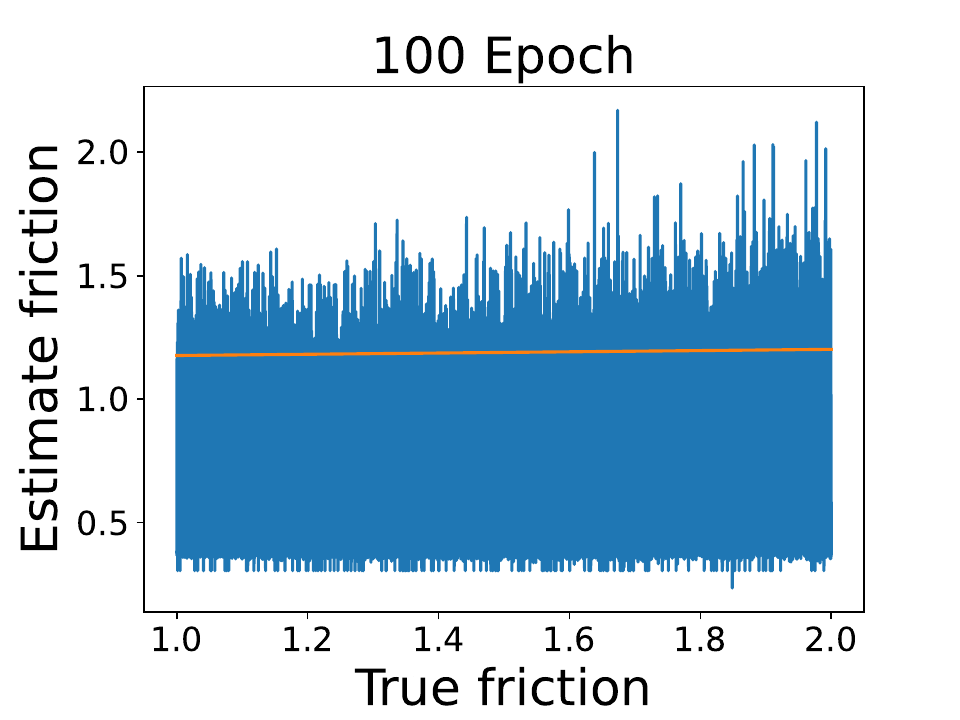} &
        \includegraphics[width=0.33\linewidth]{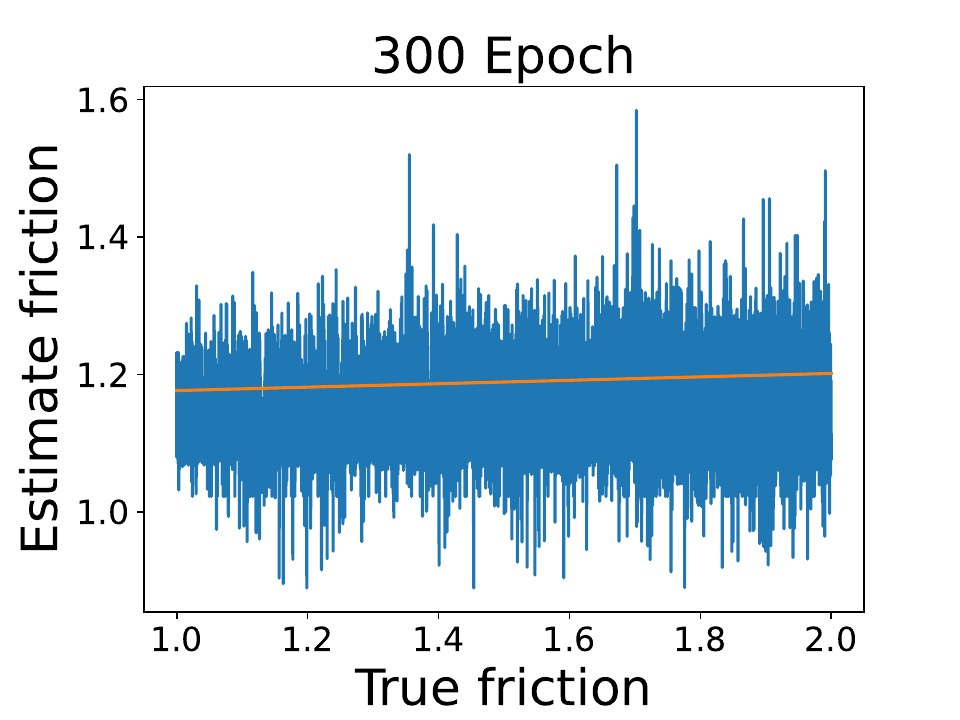} &
        \includegraphics[width=0.33\linewidth]{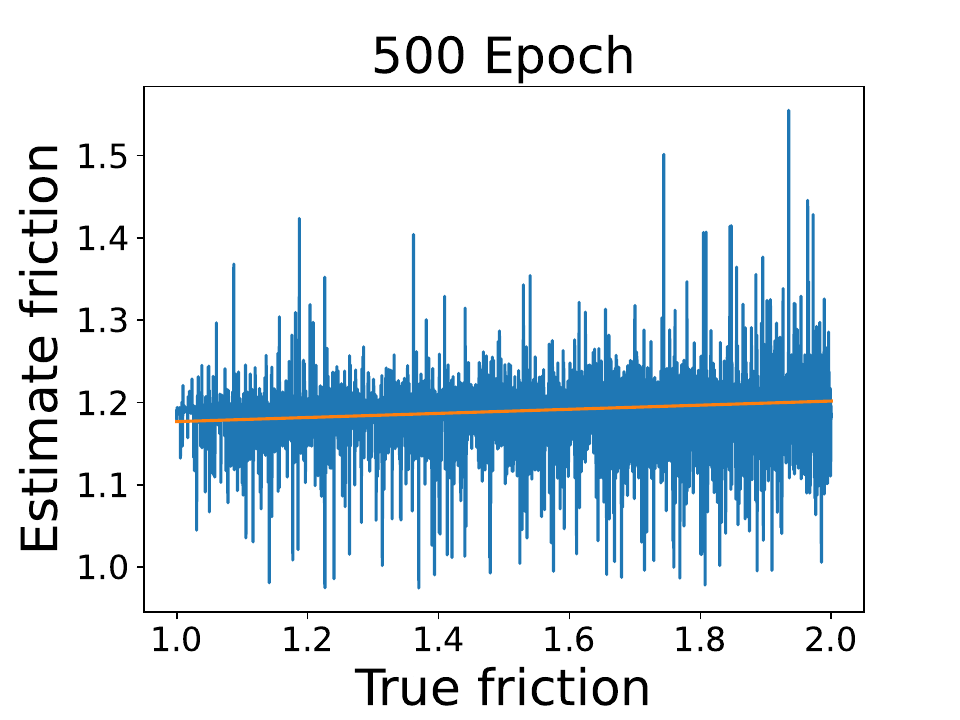} \\
    \end{tabular}
    \caption{\textbf{Estimated friction from the Adaptation Module in the RMA.} We observed that while the Adaptation Module's output closely follows the pattern of the ground truth friction, it is quite noisy. Originally, the Adaptation Module was not designed to estimate true parameters. This implies that using a deterministic value for the estimated friction is more effective in enhancing the stability and tracking performance of the wheeled robot compared to using a latent vector.}
    \label{fig:training process estimation}
     \vspace{-1.5em}
\end{figure}

%% file: 5_results_and_analysis.tex
\section{Results and Analysis} 
\label{sec:results and analysis}
\subsection{Friction Coefficient Estimation Results} 
The friction coefficient estimation results are summarized in Fig. \ref{fig:friction estimation result}. Despite the FFV module with VLM performs a bit worse than a Vision Transformer (ViT) and Word MM, our FFV module provides reasonably accurate values, particularly for materials that are known to be more slippery or rough (see Fig. \ref{fig:friction estimation result}-C). Moreover, it is quite obvious that ViT module tends to overfit due to the limited size of the training dataset.

We observed that FFV module has ability to handle edge cases involving additional material properties, we altered surface conditions by adding substances like water or oil and evaluated each model’s response. As shown in Fig.~\ref{fig:cof_addition_result}, ViT fails to respond appropriately—for example, it predicts a higher CoF for an oil-covered surface than for a dry one. In contrast, the VLM, leveraging the RAG approach, adjusts its estimates more reasonably, suggesting that it can better incorporate contextual cues such as surface material or added substances. WordMM assumes perfect surface-type classification, which explains its relatively strong performance under these conditions.

\subsection{Tracking Performance Comparison in Simulation} 
We report the simulation tracking performance comparison results in Fig. \ref{fig:tracking_error_comparison}. Our method attained the smallest tracking error, even outperforming the teacher policy slightly. This suggests that incorporating the friction coefficient into an RL policy is more effective than relying on the encoder network's latent vector. A similar finding is reported in \cite{ji2022concurrent}, where the estimator is trained directly rather than using a latent vector. The student policy shows significantly higher tracking error variability compared to other methods. We observed that while the adaptation module can reproduce the encoder network using only the robot's proprioception history, the system tends to fall suddenly and cannot recover due to its limited contact points with the ground. This suggests that relying solely on proprioception may not be ideal for learning recovery motions (e.g., RMA) in systems with few degrees of freedom and contact points. Using a DR can enhance the robustness of a RL policy, but  its overall effectiveness is constrained, offering only marginal improvements. Furthermore, we observed that incorporating a translational joint significantly enhances tracking performance. This is attributed to the fact that relying solely on the wheel joint does not provide sufficient information to accurately capture slip behavior. In the case of LQR, the lack of knowledge about ground friction causes the system to focus solely on minimizing tracking error, which often leads to failure.

\begin{figure*}[t]
    \centering
    \begin{subfigure}[b]{0.7\linewidth}
        \includegraphics[width=\columnwidth]{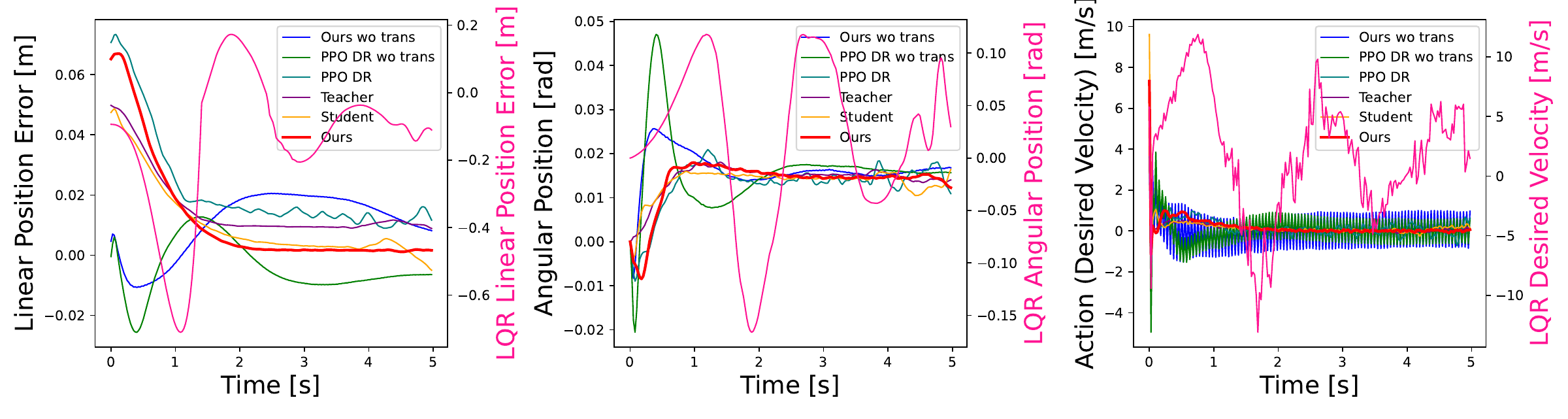}
            \caption{Trajectory Error of Each Methods in Simulation}
    \end{subfigure} 
        \hspace{0.05cm}
    \begin{subfigure}[b]{0.23\linewidth}
        \includegraphics[width=\columnwidth]{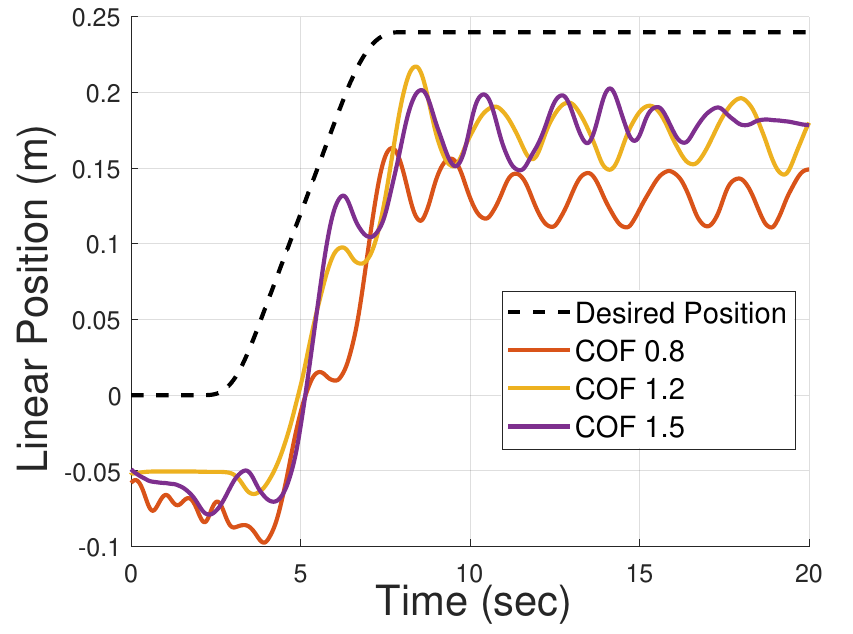}
        \caption{Performance Difference depending on Input COF in Real World}
    \end{subfigure}
\caption{\textbf{Trajectory Error for Each Method in Simulation and Performance Difference Depending on Input COF in the Real World}. (a) The trajectories represent the mean of successful samples from simulation experiments. Compared to other methods, our method (red) achieves the minimum tracking error and generates smooth actions (desired velocity). Producing smooth actions is crucial for successfully transferring a trained policy to a physical system (see the desired velocity graph). (b) The graph illustrates that our method responds differently based on the input COF, which is the estimated friction coefficient from the FFV module. This enables the system to adjust its motion in response to various ground surfaces before contact.}
\label{task_fig}
\vspace{-1em}
\end{figure*}

We compare all algorithms in simulation and present the resulting trajectories in Fig. \ref{task_fig} for detailed analysis. Methods that do not utilize the translation joint show relatively low initial tracking error but struggle to follow high-magnitude desired trajectories, resulting in stationary behavior and large steady-state errors. In contrast, our method produces the smoothest trajectories, suggesting better sim-to-real transferability, which is further validated in hardware experiments. Additionally, our approach achieves the highest reward in simulation, indicating that incorporating both friction and translation joint information is essential for learning smooth and adaptive tracking behavior.

\begin{figure}[t]
  \centering
\includegraphics[width=\linewidth]{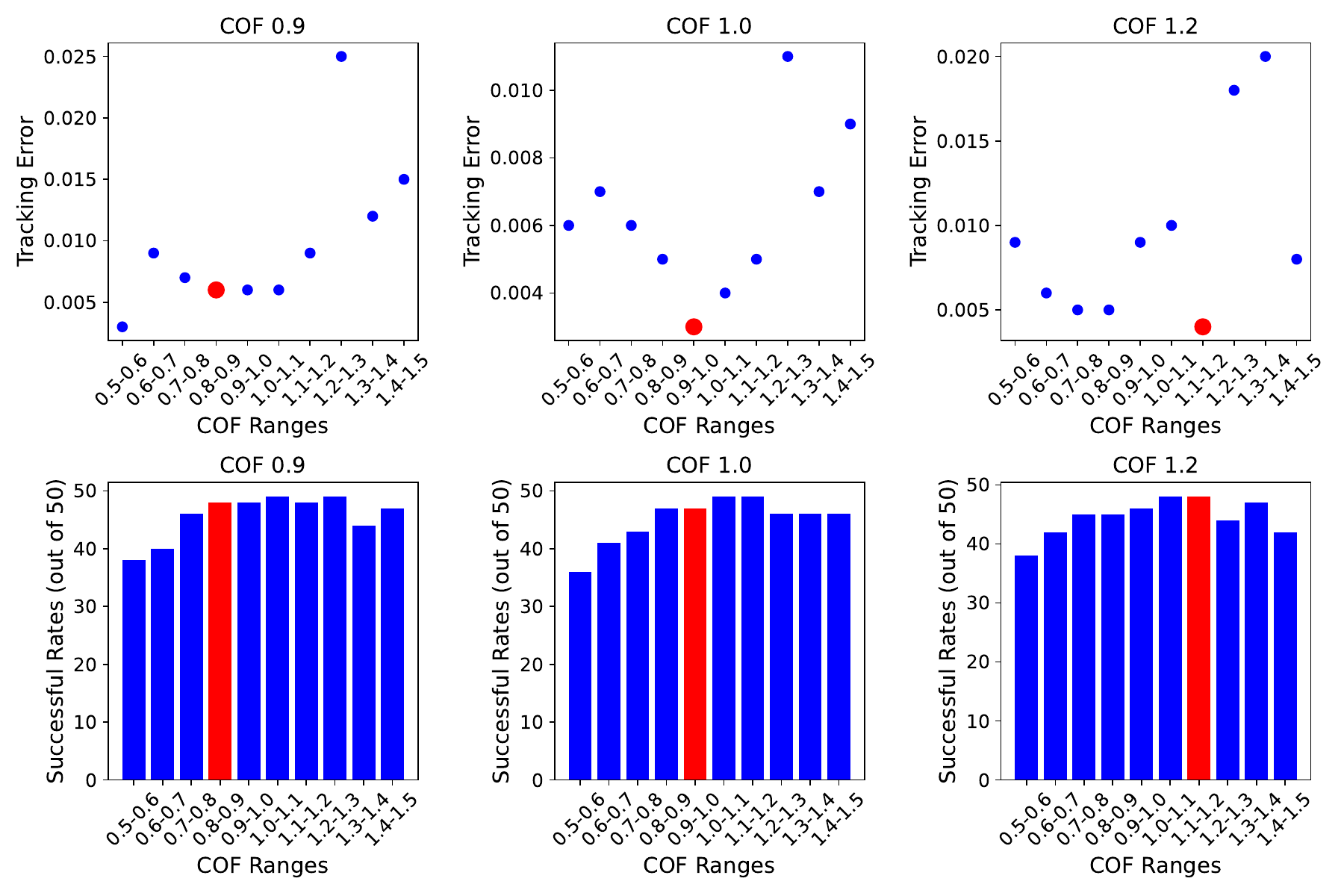}
  \caption{\textbf{Different Tracking Error and Success Rate Depending on Accuracy of Estimated COF.} This shows how inaccurate COF estimation affects the tracking performance and success rate. Better estimation performance results in lower tracking errors and higher task success rates. If the estimated COF exceeds the actual COF, the system will likely fall, resulting in a lower success rate. Conversely, tracking performance deteriorates if the actual COF is higher than estimated.}
  \label{fig:detail_cof_analysis}
  \vspace{-1em}
\end{figure}

We conducted an additional experiment to analyze the sensitivity of wheeled robot control to the accuracy of the predicted CoF. In simulation, we fixed the input CoF provided to the policy and uniformly randomized the actual environment CoF between 0.5 and 1.5. The resulting tracking error and success rate (number of successful trials out of 50) are reported in Fig. \ref{fig:detail_cof_analysis}, with corresponding real-world results shown in Fig. \ref{task_fig}.b. The simulation setup follows the same configuration described in the main manuscript.

Our method achieves almost the highest success rate and lowest tracking error when the input CoF closely matches the actual environment CoF. When the predicted CoF is lower than the ground truth, the policy behaves conservatively, which increases stability and success rate but may result in higher tracking error. Conversely, when the predicted CoF exceeds the actual value, the policy becomes overly aggressive, increasing the risk of slippage and decreasing the success rate. These findings highlight the importance of accurate CoF estimation for achieving both safe and precise locomotion.\\

\subsection{Tracking Performance Comparison in Physical System.} 
\begin{figure}[t]
  \centering
\includegraphics[width=0.9\linewidth]{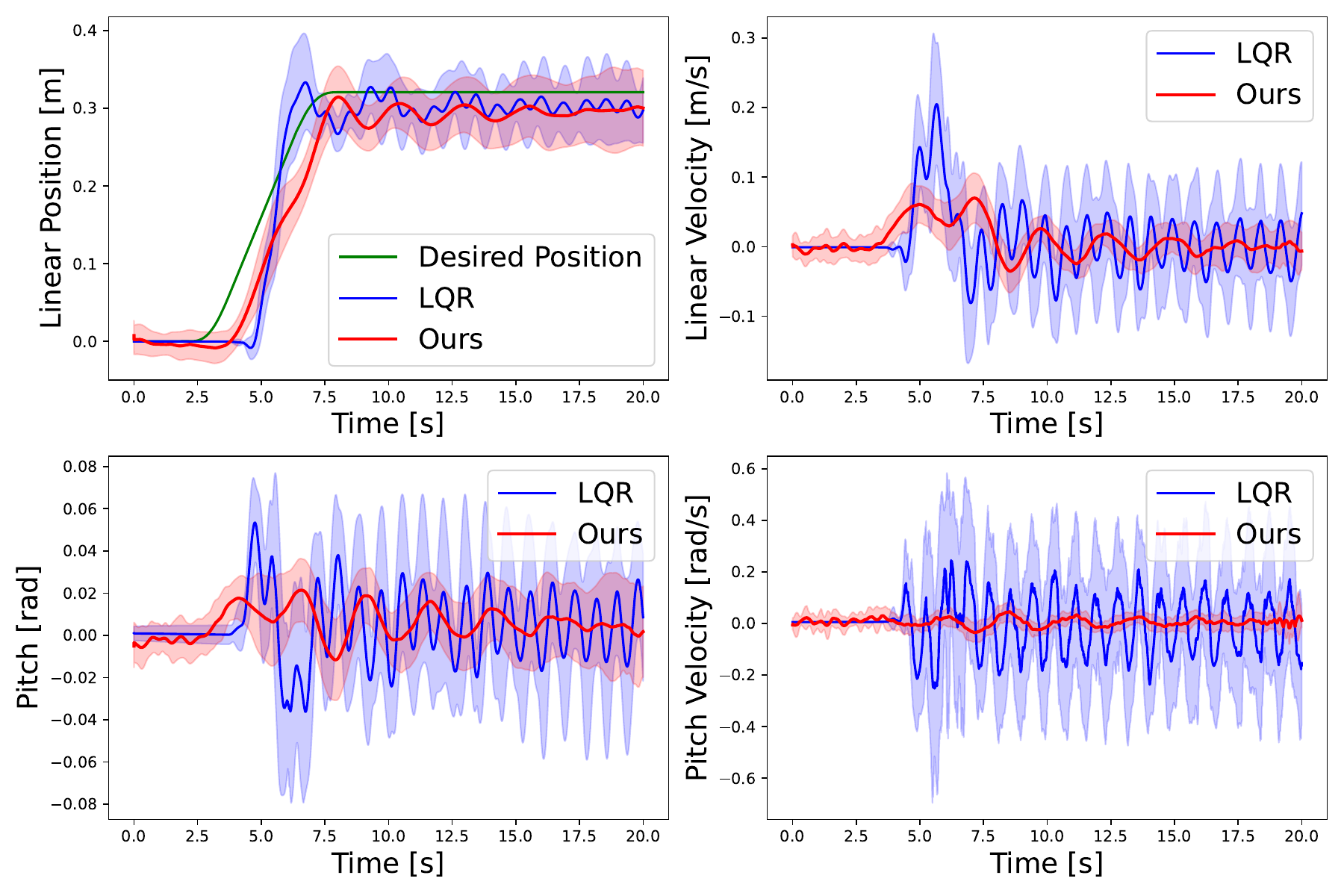}
  \caption{\textbf{Result of Tracking Task in Real World.} We evaluated our framework across combinations of five different ground surfaces and two types of wheel materials in a custom WIP system. The curves show the mean trajectories of only successful cases where the WIP remained stable, with colored areas indicating the standard deviation. Our method completed tracking tasks on all ground surfaces, whereas LQR failed on the most slippery terrain. A smooth fifth-order polynomial trajectory was used as the desired path for the algorithm to track.}
  \label{fig:real result}
\end{figure}

The tracking performance results in the real world are presented in Fig. \ref{fig:real result}, where we observed that only our framework and LQR were successfully transferred to the physical system. Other methods, such as PPO and RMA, demonstrated either overly aggressive or conservative behaviors, resulting in oscillations or stationary states. Although LQR performas the worst in simulation, it can still be transferred to the real system because in simulation, the rigid bodies are perfect and they lose contact much easier than reality while in real system, the rubber wheel acts as a buffer to allow constant rolling contact with the ground. Meaning aggressive motions would become even more aggressive in reality than simulation. Specifically, for RMA, the output variance, as noted in the simulation, was large, often leading to more aggressive motions. In contrast, domain randomization with PPO tended to produce more conservative actions, like stationary motions, to prevent falls. For systems with slippery surfaces and few contact points, this may not be a surprising result. A good example is to compare standing on one foot versus using both hands and feet to support yourself when rollerblading on slippery terrain. This challenge also lies in how quickly we can detect slip and how fast the system can react. Proprioception, which relies on recording history data, often introduces delays in response. With only a single contact point, the system lacks additional mechanical options to stabilize the robot once slip occurs, making rapid recovery difficult. This highlights the importance of using a VLM to predict terrain slipperiness in advance. 

Compared to LQR, our framework achieves better tracking with smoother, less aggressive motions. While the LQR-controlled inverted pendulum was tuned to be robust enough to complete most tasks, it failed on the most slippery surface due to its inability to adapt to varying ground conditions. Lacking awareness of surface properties, the LQR controller attempts to follow the reference trajectory blindly, leading to contact loss and system failure under low-friction scenarios. In contrast, our framework adjusts its motion based on estimated surface conditions, enhancing robustness and stability.

\begin{table}[t]
\centering
\resizebox{1\linewidth}{!}{
\begin{tabular}{lccccccc}
\toprule
\textbf{Ground Type} & \textbf{w2 + c1} & \textbf{w2 + c2} & \textbf{w1 + c3} & \textbf{w1 + c4} & \textbf{w1 + c5} & \textbf{Overall} \\
\midrule
\textbf{LQR} & 3/3 & 3/3 & 3/3 & 3/3 & 0/2 & 12/14\\
\textbf{Ours} & 3/3 & 3/3 & 3/3 & 3/3 & \textbf{5/5} & \textbf{17/17}\\
\bottomrule
\end{tabular}}
\caption{Success Rate of Tracking Task in Real World. Varying success rates across different ground surfaces were observed: w1 (more slippery wheel), w2 (rougher wheel), c1 (rubber), c2 (anti-slip tape), c3 (grass), c4 (whiteboard), and c5 (wood). The CoF between the surfaces decreases along the columns. }
\label{table:real world tracking success rate}
\end{table}

%% file: 6_discussion_and_limitations.tex
\section{Discussion and Limitations}
\label{sec:Discussion}

While our algorithm demonstrated superior performance compared to other baselines, there are still several factors that require improvement. Firstly, while our customized WIP system is more complex than traditional designs, it remains considerably simpler than wheeled humanoids \cite{purushottam2022hands}. However, we believe that our algorithm can maintain its effectiveness and versatility, given its potential for seamless integration into various other applications. This is primarily because our vision-based FFV module can be seamlessly integrated into any reinforcement learning policy, making it adaptable across different hardware platforms. Secondly, the FFV module's estimation speed is relatively slow, primarily constrained by the processing speed of OpenAI server. This is the biggest barrier to our methodology updating ground information in real time but we believe that this can be improved in the near future. Unless the ground state changes dramatically in real-time, anticipating and adjusting behavior based on the expected ground state where the robot is going to traverse on a few seconds later is enough. Estimating the friction of the surface that humans will be on a few seconds later is sufficient. One interesting avenue for future work could involve applying our algorithm to tele-operation scenarios \cite{baek2023study}, aiming to mitigate human errors by dynamically adjusting the user's commands to accomplish tasks more safely. Thirdly, the absence of a large-scale friction dataset limits the accuracy of friction estimation. While VLMs combined with the RAG technique can produce reasonable predictions, there is significant potential for improvement through the use of large datasets to fine-tune or pretrain a foundation model specifically for physical parameter estimation.

Regarding sim-to-real transfer, we identified several key factors that significantly affect success. First, the presence of additional passive mechanisms, such as roll and translation joints—which are not present in typical mobile wheeled-legged robots—makes transfer more difficult. These passive joints introduce high, unmodeled friction and cannot be directly controlled in either simulation or the real world. As a result, the system frequently exhibits point contact between the wheels and the ground, which rigid-body simulators struggle to handle accurately. Second, smooth action outputs are critical. While some policies may achieve high rewards in simulation, they often exhibit bang-bang behavior with high-frequency switching, which is unrealistic and can be harmful to physical hardware. Ensuring smoother control trajectories improves robustness and safety during real-world execution. Third, using a compact observation space improves transferability. Larger observation spaces introduce more variability and potential overfitting to simulation-specific features, increasing the likelihood of unexpected behavior in the real world. 

%% file: 7_conclusion.tex
\section{Conclusion}
\label{sec:Conclusion}
In this work, friction-aware safety locomotion framework using vision language models is proposed. We first introduce the FFV module with VLMs, which can estimate the ground friction coefficient—a parameter that is difficult to measure and for which there is limited dataset availability. Although it still has limitations in real-time performance, it can predict the risk of the system falling in advance and further improve driving safety by reasonably understanding the ground condition by utilizing texture information, thereby successfully completing driving tasks. We demonstrate both in simulation and in the real world, showing that it enables adaptive behavior in wheeled robots and helps mitigate slip risks across various ground surfaces. Our implementation is straightforward and integrates seamlessly with existing PPO-based methods without requiring structural changes. We believe our work opens up intriguing research directions, such as exploring whether a VLM can be used to assess physics parameters beyond just friction.